\def\year{2020}\relax

\documentclass[letterpaper]{article} 
\usepackage{aaai20}  
\usepackage{times}  
\usepackage{helvet} 
\usepackage{courier}  
\usepackage[hyphens]{url}  
\usepackage{graphicx} 
\usepackage{graphicx}  
\usepackage{booktabs, adjustbox} 
\usepackage{amsmath, amssymb}
\usepackage{multirow}

\usepackage{epstopdf}
\usepackage[caption=true,font=footnotesize]{subfig}

\title{Adaptive Anomaly Detection for IoT Data in Hierarchical Edge Computing}
\author{Mao~V.~Ngo,\textsuperscript{\rm a,\rm b\thanks{This work was supported in part by the CNRS Grant SAMOVAR DAD4IoT 2019.}}
     Hakima~Chaouchi,\textsuperscript{\rm c}
     Tie~Luo,\textsuperscript{\rm d}
     Tony~Q.S.~Quek\textsuperscript{\rm a}
     \\ 
\textsuperscript{\rm a}Singapore University of Technology and Design, Singapore \quad 
\textsuperscript{\rm b}Institute for Infocomm Research, A*STAR, Singapore\\
\textsuperscript{\rm c}CNRS, SAMOVAR, Telecom Sud Paris, Institut Mines-Telecom, Paris-Saclay University, France\\
\textsuperscript{\rm d}Department of Computer Science, Missouri University of Science and Technology, USA\\
vanmao\_ngo@mymail.sutd.edu.sg,
hakima.chaouchi@telecom-sudparis.eu,
tluo@mst.edu,
tonyquek@sutd.edu.sg
}

\usepackage{color}

\makeatletter
\def\thanks#1{\protected@xdef\@thanks{\@thanks
	\protect\footnotetext{#1}}}
\makeatother
\begin{document}


\maketitle

\begin{abstract}
Advances in deep neural networks (DNN) greatly bolster real-time detection of anomalous IoT data. 
However, IoT devices can barely afford complex DNN models due to limited computational power and energy supply.
While one can offload anomaly detection tasks to the cloud, it incurs long delay and requires large bandwidth when thousands of IoT devices stream data to the cloud concurrently.
In this paper, we propose an adaptive anomaly detection approach for hierarchical edge computing (HEC) systems to solve this problem. 
Specifically, we first construct three anomaly detection DNN models of increasing complexity, and associate them with the three layers of HEC from bottom to top, i.e., IoT devices, edge servers, and cloud. Then, we design an adaptive scheme to select one of the models based on the contextual information extracted from input data, to perform anomaly detection. The selection is formulated as a {\em contextual bandit problem} and is characterized by a single-step Markov decision process, with an objective of achieving high detection accuracy and low detection delay simultaneously. 
We evaluate our proposed approach using a real IoT dataset, and demonstrate that it reduces detection delay by 84\% while maintaining almost the same accuracy as compared to offloading detection tasks to the cloud. In addition, our evaluation also shows that it outperforms other baseline schemes.
\end{abstract}


\section{Introduction}
\label{sec:introduction}

With the increasing demand of detecting anomalous sensory data generated by a massive number of IoT devices, machine learning---especially deep learning---offers an effective approach and has been successfully applied to many anomaly detection tasks in IoT environments~\cite{Mohammadi_SurveyIoTBigData_2018,Luo_ICC2018,Malhotra_LSTM_encDec_ICMLWrsh2016}.
A variety of IoT applications, such as collision avoidance for autonomous vehicles and fire alarm system in factories, are time-critical and requires fast anomaly detection. In these cases, the traditional approach of streaming all the IoT sensory data to the cloud can be problematic as it tends to incur high communication delay, congest backbone network, and pose a risk on data privacy. 

Anomaly detection with edge or fog computing \cite{La_FogComputing_2019,Chen_SEC2017} can provide a better alternative by performing distributed anomaly detection in the proximity of the sources of sensory data. However, pushing computation from cloud to the edge faces resource challenges especially when the model is complex (such as deep learning models) and the edge device only has limited computation power, storage, and energy supply, which is the case for typical IoT devices.

A possible solution is to transform a large complex model into one that fits the IoT device's capability; for example, model compression \cite{Han2015DeepCompression} achieves this by pruning redundant and unimportant (near-zero) parameters as well as by quantizing weights into bins. However, such an approach needs to handle the anomaly detection models on a case-by-case basis via fine-tuning, and is only applicable to a few specific types of deep neural networks (DNNs) with large sparsity.

There are also other proposed approaches \cite{Teerapittayanon_ICDCS2017,Neurosurgeon_Kang2017}, but overall, we identify three main issues in most of existing works:
(1) ``one size fits all'' - prior work attempts to use one anomaly detection model to handle all input data, while it is generally overlooked that different data samples often have different levels of hardness in detecting anomaly events;
(2) improving accuracy or F1-score was often the focus but detection delay and memory footprint were not given adequate consideration;
(3) lack of appropriate local analysis and thus data are often transmitted back and forth between sources and the cloud, incurring unnecessary delay.


In this paper, we propose an adaptive approach that leverages the hierarchical edge computing (HEC) system by adaptively matching data of differentiated hardness level for detection with models of different complexity. Specifically, we construct three anomaly detection DNN models of increasing complexity based on the state of the art, and associate them with the three layers of HEC from bottom to top, i.e., IoT devices, edge servers, and cloud. Then, we propose an adaptive scheme that judiciously selects one of the models based on the contextual information extracted from the particular input data to perform anomaly detection. The scheme follows a single-step Markov decision process by formulating the model selection problem as a {\em contextual bandit problem}.
By selecting appropriate models, we avoid unnecessary data transmission to edge servers and the cloud, and minimize detection delay as well.



Our contributions are summarized as follows:
\begin{itemize}
    \item We identify three main issues in existing IoT anomaly detection approaches, namely using one universal model to fit all scenarios, one-sided focus on accuracy, and lack of local analysis that results in unnecessary traffic.
    
    \item We propose an anomaly detection approach that consists of two components: (i) constructing three DNN models based on state of the art with increasing complexity, and associating them with three corresponding layers of HEC systems; (ii) designing an adaptive scheme to select appropriate models based on the contextual information of input data, by casting the problem as a contextual bandit problem which is also represented as a single-step Markov decision process.
    
    \item We evaluate our approach using a real-world IoT dataset and show that it outperforms other benchmark schemes; in particular, it makes the best tradeoff by achieving high detection accuracy and low detection delay simultaneously.
    
\end{itemize}

\section{Related work}
\label{sec:relatedWork}
Deep learning is becoming increasingly popular in anomaly detection for IoT applications \cite{Mohammadi_SurveyIoTBigData_2018,Luo_ICC2018,singh2017anomaly,Malhotra_LSTM_encDec_ICMLWrsh2016}. 
For example, \cite{Luo_ICC2018} proposed an autoencoder (AE) neural network model for distributed anomaly detection in IoT systems. They built an AE neural network to capture the patterns of normal data, and detect outliers based on {\em reconstruction errors}. Their AE model can be deployed at IoT devices to perform distributed anomaly detection, but the model is lightweight and may not be able to detect some complex anomalous events.
A more complex model was proposed by \cite{Malhotra_LSTM_encDec_ICMLWrsh2016} which used an LSTM-based encoder-decoder model to predict events in a few future time-steps based on a few historical time-steps. 
However, that model does not suit resource-constrained IoT devices due to its high computational cost.

On another line of research, 
\cite{Teerapittayanon_ICPR2016} proposed a BranchyNet architecture for an image classification task that can early ``exit'' from a multi-layer DNN during inference based on the confidence of inference output.
Later on, the same authors \cite{Teerapittayanon_ICDCS2017} deployed different sections of BranchyNet in a HEC system, in order to reuse extracted features from lower layers to do inference at a higher layer. This requires less communication bandwidth and allows for faster and localized inference due to the shallow model at the edge.
However, this approach has to make inference sequentially from the very bottom to the top of HEC, which can lead to unnecessary delay and inference requests to lower layers when detection is hard. In addition, it requires all the distributed models to use the same architecture, while our approach has the flexibility of using different models at different layers of HEC.

Our work was inspired by \cite{Taylor_SIGPLAN2018} and \cite{Blockdrop_WuCVPR2018} from the computer vision domain for image classification applications. \cite{Taylor_SIGPLAN2018} used multiple k-Nearest Neighbor (KNN) classification models to train a selection model to choose a proper inference model (among several models within an embedded device) for a given input image and desired accuracy. \cite{Blockdrop_WuCVPR2018} proposed a Blockdrop scheme that learns to dynamically drop or keep a residual block of a trained deep residual networks (ResNets) during inference, in order to minimize the number of residual blocks and preserve recognition accuracy. 
In this paper, we introduce a similar idea into the IoT domain and apply it to anomaly detection tasks. Furthermore, instead of using multiple sequentially KNN classifiers like \cite{Taylor_SIGPLAN2018} which is hard to scale, we design a {\em policy network} that can directly outputs a suitable model (rather than sequentially checking each KNN classifier until the required accuracy is met) based on the contextual information of input IoT data.

\section{Adaptive Anomaly Detection}
\label{sec:AdaptiveAnomalyDetection}

We consider a $K$-layer HEC system: IoT devices at layer-1, edge servers at layer-2 to layer-$(K-1)$, and the cloud at layer-$K$. In Section~\ref{subsec:multipleADModels}, we construct $K$ anomaly detection models with increasing complexity and associate them with the layers from 1 to $K$. In Section~\ref{subsec:DynamicModelSelectionScheme}, we design an adaptive model selection scheme
that automatically chooses a suitable model to detect anomaly based on input data, in order to achieve high accuracy and low detection delay simultaneously, and in the meantime consumes less uplink bandwidth.


\subsection{Multiple Anomaly Detector Models in Hierarchical Edge Computing}
\label{subsec:multipleADModels}
\begin{figure}[t]
    \centering
    \includegraphics[width=1.0\linewidth]{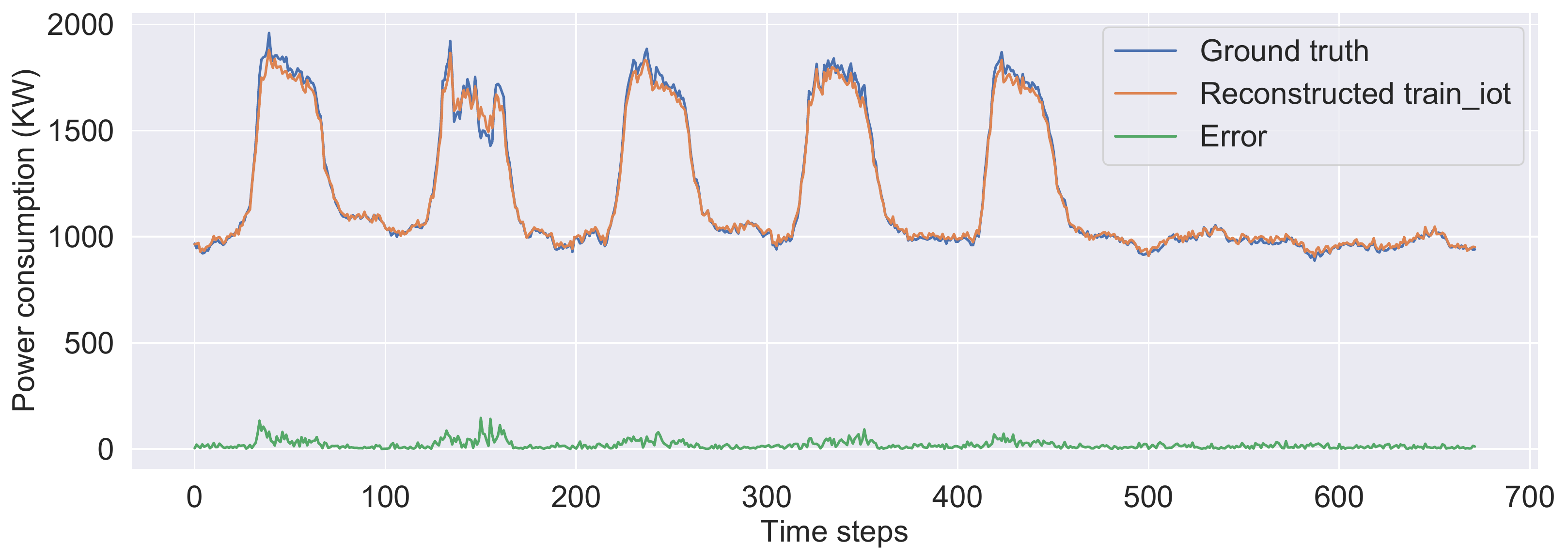}
    \caption{Reconstruction error of AE-IoT model.}
    \label{fig:ReconstructionTrainAutoencoderIoT}
\end{figure}

As is common in the literature \cite{La_FogComputing_2019,Mohammadi_SurveyIoTBigData_2018}, we consider $K=3$ for HEC, corresponding to IoT devices, edge servers, and the cloud.
For these three layers from bottom to top, we construct three anomaly detection models of increasing complexity. 
For IoT devices, we adapt the autoencoder (AE) model with a single hidden layer from \cite{Luo_ICC2018}, which has proved the feasibility of running this model on IoT devices. 
In \cite{Luo_ICC2018}, the compression ratio between the dimension of the encoded layer and that of the input layer is 70\%; while in our case, we use a much lower ratio of 30\% in order to fit more diverse low-cost IoT devices. Simulation shows that our model under this ratio can still reconstruct normal data very well (see Fig.~\ref{fig:ReconstructionTrainAutoencoderIoT}).

For edge servers (e.g., IoT gateways or micro-servers), we have a wider choice from more complex state-of-the-art anomaly detection models such as deeper AE, LSTM \cite{singh2017anomaly,Malhotra_LSTM_encDec_ICMLWrsh2016}, and generative adversarial networks (GAN). In this study, we consistently use AE-based model with more hidden layers to enhance the capability of learning better features to represent data. We add one more encoder layer and one more decoder layer to the previous AE model to obtain a model which we call \textit{AE-Edge}. For the cloud, we further add one more encoder layer and decoder layer to have a deep AE model, which we refer to as \textit{AE-Cloud}.

The detailed setup of the above models is given in Table~\ref{tab:threeAnomalyDetectors}. Each number in the ``Architecture'' column is the number of units of a corresponding layer. 
The column ``\# of Parameters'' (weights and biases) reflects the approximate {\em memory footprint} of each model.
The ``FLOP'' column gives the total number of floating point operations, which reflects the required computation of each model during the inference phase.

\begin{table}[t]     
    \caption{Three anomaly detection models for IoT devices, edge severs, and the cloud.}
    \label{tab:threeAnomalyDetectors}
    \centering 
    \scriptsize
    \begin{tabular}{ l@{\hspace{0.5em}} l@{\hspace{0.5em}} l@{\hspace{0.5em}} l@{\hspace{0.5em}} }
        \toprule
         \textbf{Model}    & \textbf{Architecture} & \textbf{\# of Parameters} & \textbf{FLOP} \\
        \hline%
        \multirow{1}{*}{\bf AE-IoT}  & Input(672), Hidden(201), Output(672)  & \multirow{1}{*}{271,017} & \multirow{1}{*}{1.35 Million}\\
        \hline%
        \multirow{2}{*}{\bf AE-Edge} & Input(672), Hidden(336 \textbar 201 \textbar 336),  &  \multirow{2}{*}{949,468} & \multirow{2}{*}{2.93 Million} \\
                                    & Output(672) & & \\
        \hline%
        \multirow{2}{*}{\bf AE-Cloud} & Input(672), Hidden(470 \textbar 336 \textbar 201\textbar & \multirow{2}{*}{1,085,077} &  \multirow{2}{*}{5.41 Million} \\
                                    & \textbar 336 \textbar 470), Output(672) & & \\
        \bottomrule
    \end{tabular}		
\end{table}

We train our models with stochastic gradient descent (SGD) to minimize the mean absolute error between the reconstructed outputs and the expected outputs (which are equal to the inputs).
To avoid overfitting, we use $\ell_2$-norm regularization for weights, and add a dropout rate of 0.3 after each hidden layer.
In accordance with the different complexities of these models, we train them over 4000, 6000 and 8000 training epochs for AE-IoT, AE-Edge, and AE-Cloud, respectively.

The training process shows that the above AE models can well capture the normal data pattern, indicated by low reconstruction error for normal data and high error for abnormal data (the reconstruction error is $e_i=|x_i-\hat{x_i}|$ where $x_i$ is input data and $\hat{x_i}$ is the corresponding reconstructed output). Therefore, the reconstruction error is a good indicator for detecting anomaly. 
Assuming that reconstruction errors generally follow Gaussian distribution $\mathcal N(\mathbf{\mu}, \mathbf{\Sigma})$, where $\mathbf\mu$ and $\mathbf{\Sigma}$ are the mean and covariance matrix of reconstruction errors of normal data samples (the training set), we use \textit{logarithmic probability densities (logPD)} of the reconstruction errors as \textit{anomaly scores}, as is similar to \cite{singh2017anomaly,Malhotra_LSTM_AD_ESANN2015}. 
The normal data will have a high logPD while anomalous data will have a low logPD.
We then use the minimum value of the logPD on the normal dataset (i.e., the training set) as the threshold for detecting outliers during testing.

\subsection{Adaptive Model Selection Scheme}
\label{subsec:DynamicModelSelectionScheme}

As AE-IoT, AE-Edge, and AE-Cloud will be deployed in a HEC corresponding to the three layers, we propose an adaptive model selection scheme in this section to select the most suitable model based on the contextual information of input data (testing sample), so that each data sample will be directly fed to its best-suited model. Note that this is in contrast to traditional approaches where input data will either (i) always go to one same model regardless of the hardness of detection \cite{Chen_SEC2017}, or (ii) be successively offloaded to higher layers until meeting a required or desired accuracy or confidence \cite{Teerapittayanon_ICDCS2017}. 

Our proposed adaptive model selection scheme is a reinforcement learning algorithm that adapts its model selection strategy to maximize the expected reward of the model to be selected.
We frame the learning problem as a \textit{contextual bandit problem} \cite{sutton2000policy,williams1992REINFORCE} (which is also known as associative reinforcement learning (RL), one-step RL, associative bandits, and learning with bandit feedback) and use a {\em policy gradient method} to solve it. 

\begin{figure}[t]
    \centering
    \includegraphics[width=1.0\linewidth]{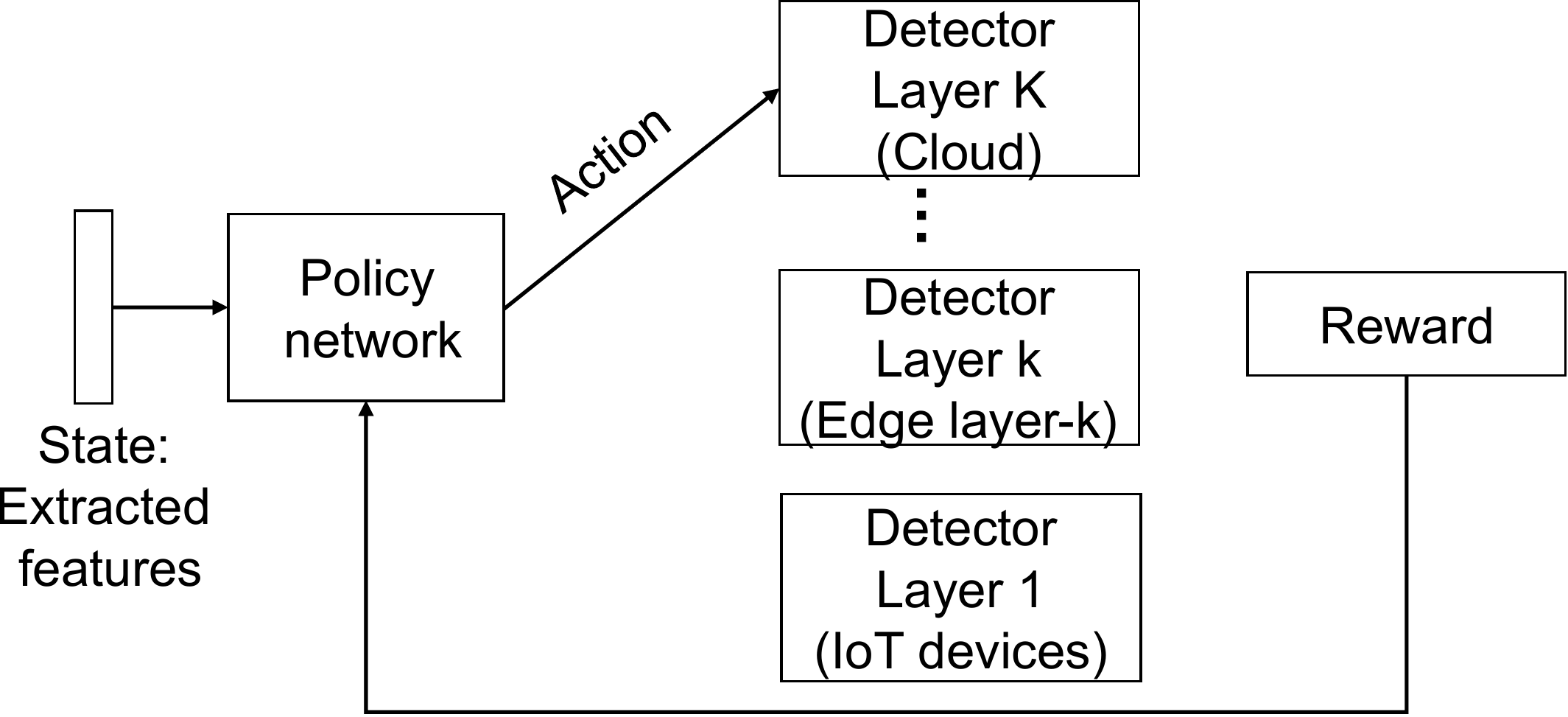}
    \caption{Adaptive model selection based on a policy network and modeled as a contextual bandit problem.}
    \label{fig:DecisionMakingModule}
\end{figure}

See Fig.~\ref{fig:DecisionMakingModule}. Formally, given the contextual information $\mathbf{z_x}$ of an input data $\mathbf{x}$, where $\mathbf{z_x}$ is composed of extracted features of the input data $\mathbf{x}$, 
and $K$ trained models deployed at the $K$ layers of a HEC system,
we build a policy network that takes $\mathbf{z_x}$ as the input state and outputs a policy of selecting which model (or equivalently which layer of HEC) to do anomaly detection, in the form of a categorical distribution
\[ \pi_{\theta} (\mathbf{a}|\mathbf{z_x}) = \prod_{k=1}^{K} s_k^{a_k} ,\]
where $\mathbf{a}=(a_1, a_2,\cdots,a_K)$, $a_k\in\{0,1\}$, is the actions encoded as a one-hot vector which defines which model (or HEC layer) to perform the task, $\mathbf{s}=(s_1, s_2,\cdots, s_K)=f_{\theta}(\mathbf{z_x})$, $s_k \in [0,1]$, is a likelihood vector representing the likelihood of selecting each model $k$, and $\sum_{k=1}^{K} a_k = 1, \sum_{k=1}^{K} s_k = 1$. We set $a_k=1$ if $k=\arg \max_{k} (s_k)$ and $a_k=0$ otherwise, and we denote the selected action as $|\mathbf{a}|=k$.

The policy network $f_{\theta}(.)$ is designed as a single hidden layer neural network with parameters $\theta$. To make the policy network small enough to run fast on IoT devices, we use the extracted features $\mathbf{z_x}$ instead of the raw input data $\mathbf{x}$, to represent the contextual information of input data (i.e., a state vector). 
The policy network is trained to find an optimal policy $\pi$ that maps an input state $\mathbf{z_x}$ to an action (which model or layer) to maximize the expected reward of the selected actions.
We train the policy network using a {\em policy gradient method} \cite{williams1992REINFORCE,sutton2000policy} to minimize the negative expected reward:
\[ 
\min L(\theta) = -\mathop{\mathbb{E}}_{\mathbf{a} \sim \pi_{\theta}}[R(\mathbf{a},\mathbf{z_x})],
\]
where $R(\mathbf{a},\mathbf{z_x})$ is a reward function of action $\mathbf{a}$ given state $\mathbf{z_x}$.

To reduce the variance of reward value and increase the convergence rate, we utilize {\em reinforcement comparison} \cite{williams1992REINFORCE} with a baseline $R(\tilde{\mathbf{a}}, \mathbf{z_x})$ that is independent of output actions. 
We choose the baseline $R(\tilde{\mathbf{a}}, \mathbf{z_x})$ to be the best observed reward \cite{sutton2000policy}, which is empirically shown to boost the convergence rate.
In addition, we add a $\ell_2$-norm regularization term to prevent overfitting.
So the loss function $L(\theta)$ is rewritten as follows: 
\begin{align}
L(\theta) =  -\mathop{\mathbb{E}}_{\mathbf{a} \sim \pi_{\theta}}[(R(\mathbf{a},\mathbf{z_x})-R(\tilde{\mathbf{a}}, \mathbf{z_x})] + \frac{\gamma}{2} ||\theta ||_2,
\label{eq:policyGradientRegularization}
\end{align}
where $\gamma$ is a regularization parameter. 
To minimize \eqref{eq:policyGradientRegularization}, we utilize the policy gradient with REINFORCE algorithm \cite{williams1992REINFORCE} to compute the gradient of $L(\theta)$ as follows:
\begin{align}
\fontsize{6.5pt}{10.0pt}\selectfont
&\nabla_{\theta} L \notag \\
& =  -\mathop{\mathbb{E}}_{\mathbf{a} \sim \pi_{\theta}} \left[ (R(\mathbf{a},\mathbf{z_x})-R(\tilde{\mathbf{a}}, \mathbf{z_x}))\nabla_{\theta} \log(\pi_{\theta}(\mathbf{a}|\mathbf{z_x}) ) \right] + \gamma \theta \notag \\
& = - \mathop{\mathbb{E}}_{\mathbf{a} \sim \pi_{\theta}} \left[ (R(\mathbf{a},\mathbf{z_x})-R(\tilde{\mathbf{a}}, \mathbf{z_x}))\nabla_{\theta} \sum_{k=1}^{K} a_k \log(s_k) \right] + \gamma \theta \notag 
\end{align}

In order to encourage selecting an appropriate model that jointly increases accuracy and reduces the cost of offloading tasks further away from the edge, we propose a reward function as follows:
\[
R(\mathbf{a}, \mathbf{z_x}) = \text{accuracy}(\mathbf{x}) -  C(\mathbf{a}, \mathbf{x}),
\]
where accuracy($\mathbf{x}$) is the accuracy of detecting anomaly in an input $\mathbf{x}$, 
and $C(\mathbf{a},\mathbf{x})$ is the cost function of offloading the detection task to a layer $k=|\mathbf{a}|$ for data $\mathbf{x}$, which will be defined later.

To balance between exploration and exploitation during training, we apply a simple $\epsilon$-greedy approach for action selection. 
With probability $\epsilon$, an action is randomly selected for exploration of more actions, while with probability $(1-\epsilon)$, an action is greedily selected based on output of the current policy network.
In implementation, we gradually decrease the value of $\epsilon$ after a certain number of episodes, so as to exploit more of the current policy network after it is sufficiently trained.

We define the cost function $C(\mathbf{a}, \mathbf{x})$ as a function of total delay, which consists of the communication delay 
$t_\text{commun}(\mathbf{x}, \mathbf{a})$ of transmitting data $\mathbf{x}$ from an IoT device to a server at layer $k=|\mathbf{a}|$ of HEC,
and the computation delay $t_\text{comp}(\mathbf{a})$ of executing the detection task at layer $k=|\mathbf{a}|$:
\[
C(\mathbf{a}, \mathbf{x}) = f_\text{cost}(t_\text{commun}(\mathbf{x}, \mathbf{a}) + t_\text{comp}(\mathbf{a}))
\]
The computation delay is given by
\begin{align}
\label{eq:ComputeDelay}
t_\text{comp}(\mathbf{a})=
    \frac{\text{FLOP of model } k=|\mathbf{a}|}{\text{FLOPS of machine } k=|\mathbf{a}|},
\end{align}
where the total FLOP (floating point operations) of a model $k$ can be measured during training for just once, as shown in the last column of Table \ref{tab:threeAnomalyDetectors}.
The FLOPS (FLOP per second) of a machine $k$ can be measured by a benchmark tool \texttt{linpack}\footnote{http://www.netlib.org/benchmark/hpl/}.

The function $f_\text{cost}(t)$ maps the total delay $t$ to an equivalent accuracy in scale $[0,1]$.
For example, $f_\text{cost}(50\,\text{ms})=0.1$ means the offloading cost 50\,ms of delay is equivalent to a reduction 0.1 in accuracy.
We use the following function as the mapping function where a higher delay will result in a greater reduction in accuracy:
\begin{align}
\label{eq:costFunction}
f_\text{cost}(t) =  \frac{\alpha \cdot t}{1+ \alpha \cdot t},
\end{align}
where $\alpha$ is a tunable parameter.

\section{Performance Evaluation}
\label{sec:ExpSetup}
\subsection{Dataset}
\begin{figure}[htb]
    \centering
    \subfloat[Normal week]{
        \includegraphics[width=0.75\linewidth]{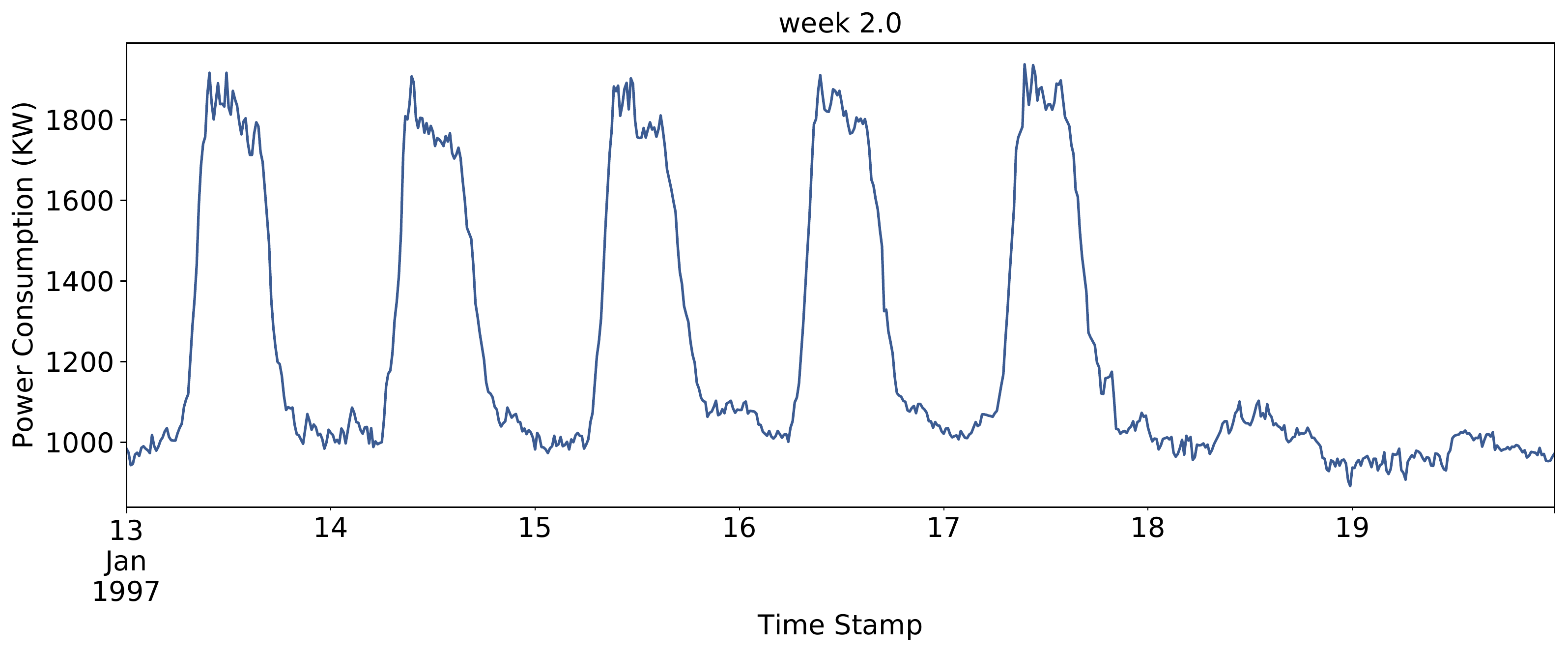}%
        \label{fig:PowerDemandNormalWeek}%
    } \vfil
    \subfloat[Abnormal week]{
        \includegraphics[width=0.75\linewidth]{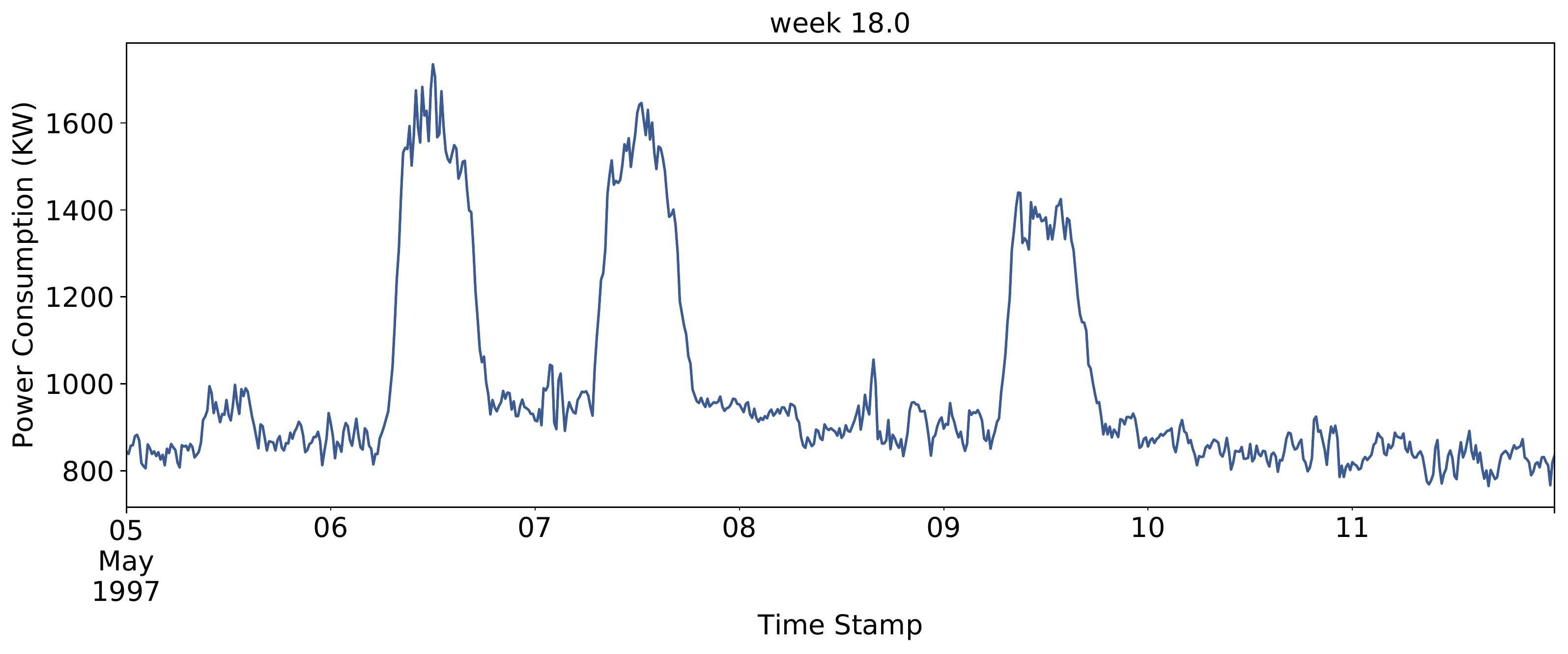}%
        \label{fig:PowerDemandAbnormalWeek}%
    }
    \caption{Example weeks of power consumption.}
    \label{fig:ExamplePowerDemand}
\end{figure}

We evaluate our proposed approach using a public dataset of power consumption\footnote{\url{http://www.cs.ucr.edu/~eamonn/discords/}} of a Dutch research facility, 
which has been used in \cite{Keogh_HotSax_ICDM2005,Malhotra_LSTM_encDec_ICMLWrsh2016,singh2017anomaly}. 
It consists of 35040 samples recorded every 15 minutes, the data has a repetition of weekly cycle of 672 time steps with five consecutive peaks for five weekdays and two lows for weekends. 
The abnormal week could have less than five consecutive peak days which might be because of a holiday in a weekday, or high power consumption in a weekend.  
Examples of normal and abnormal weeks are shown in Fig.~\ref{fig:ExamplePowerDemand}.
Hence, each input data is a sequence of one week of data with 672 time steps. 
We manually label a day as abnormal if it is a weekday with low power consumption, or it is a weekend with high power consumption; other days are labeled as normal. 
For the anomaly detection task, we split the dataset into train and test sets with ratio 70:30, or equivalently 37 weeks:15 weeks. 
The train set only contains normal weeks and the test set contains the remaining normal weeks and all the 8 anomalous weeks, each having at least one abnormal day.

To train the model selection policy network, we choose a train set that contains all the 8 abnormal weeks and 7 normal weeks, and a test set that is the whole dataset to verify the quality of the policy network.
The data is standardized to zero mean and unit variance for all of the above trainings (i.e., anomaly detection models and the policy network).

\subsection{Experiment Setup}
\label{subsec:ExperimentSetup}

We use Tensorflow to implement three anomaly detection models with network architecture described in Section \ref{subsec:multipleADModels}.
The input is a sequence of 672 time steps---a week of measured power consumption.
We use \texttt{tanh} as the non-linear activation function for all the hidden layers.
We train and test the three models separately with $5$-fold cross-validation.

We then deploy the three trained anomaly detection models at a IoT device, an edge server, and a cloud server with specifications shown in Table \ref{tab:SpecificationDevices}.
Network latency shown in Table \ref{tab:SpecificationDevices} is configured by using Linux traffic control tool, \texttt{tc}, to emulate WAN connections of HEC~\cite{Chen_SEC2017}.
The peak FLOPS of the IoT device and servers shown in Table \ref{tab:SpecificationDevices} are measured by a benchmark tool, \texttt{linpack}.
Then, the computation time of the three models are calculated by \eqref{eq:ComputeDelay}.
The total delay which includes both computation and communications will be evaluated by simulation.
The cost of executing each model on the IoT device, edge, or cloud is calculated by \eqref{eq:costFunction}, which depends on $\alpha$. 
We will evaluate with different parameters $\alpha \in [0.0001, 0.0045]$ to see the trade-off between the offloading cost and the accuracy gain of a complex model. 

\begin{table}[t]
    \centering 
    \scriptsize 
    \caption{Hardware specifications of the deployed IoT device, edge server, and cloud server in our experiment. The network latency is between the IoT device and a corresponding higher-layer server.}
    \label{tab:SpecificationDevices}
    \begin{tabular}{l@{\hspace{0.5em}} l@{\hspace{0.5em}} c@{\hspace{0.5em}} c@{\hspace{0.5em}}} 
        \toprule
        \textbf{Layer} & \textbf{Specifications} & \textbf{FLOPS} & \textbf{Latency} \\
        \midrule%
        \textbf{IoT device} & Broadcom BCM2837 1.2\,GHz, 4 cores & {194M} & 0ms\\
        \hline%
        \textbf{Edge server} & Intel i7-4790 3.60\,GHz, 4 cores, 8 threads & {197G} & 50ms \\
        \hline%
        \textbf{Cloud server} & {Intel i7-6850K 3.60\,GHz, 6 cores, 12 threads}& {289G} & 100ms \\
        \bottomrule
    \end{tabular}
\end{table}

In order to reduce complexity for the policy network, the state representation of raw input data is an extracted feature vector which includes \textit{min, max, mean}, and \textit{standard deviation} of each day's sensor data.
So the dimension of the contextual state is just 4x7=28.
We build a single hidden neural network with 100 hidden units and an output layer with 3 units which indicate the likelihood of choosing one of three detection models.
We train the policy network as described in Section \ref{subsec:DynamicModelSelectionScheme} with 6000 episodes and the initial $\epsilon=0.5$ is gradually decreased to zero after 3000 episodes.

We evaluate the performance of our adaptive model selection scheme with four other baseline schemes: 
(1) \textit{AE-IoT} which always detects on IoT devices, 
(2) \textit{AE-Edge} which always offloads detection to an edge server, 
(3) \textit{AE-Cloud} which always offloads to the cloud, and 
(4) \textit{Successive} which first executes on IoT devices and then successively offloads to a higher layer until obtaining a confident output or reaching the cloud. 
For the {\it Successive} scheme, a detected anomalous day is considered confident if the day contains at least one data point that has the logPD of error larger than the threshold by a certain factor. We experiment with threshold factors of 2 and 4, and name the corresponding schemes \textit{Successive-2} and \textit{Successive-4}, respectively.

\subsection{Experiment Results}
\label{subsec:ExperimentResults}
\begin{figure}[!htb]
    \centering
    \subfloat[Absolute reconstruction errors.]{
        \includegraphics[width=1.0\linewidth]{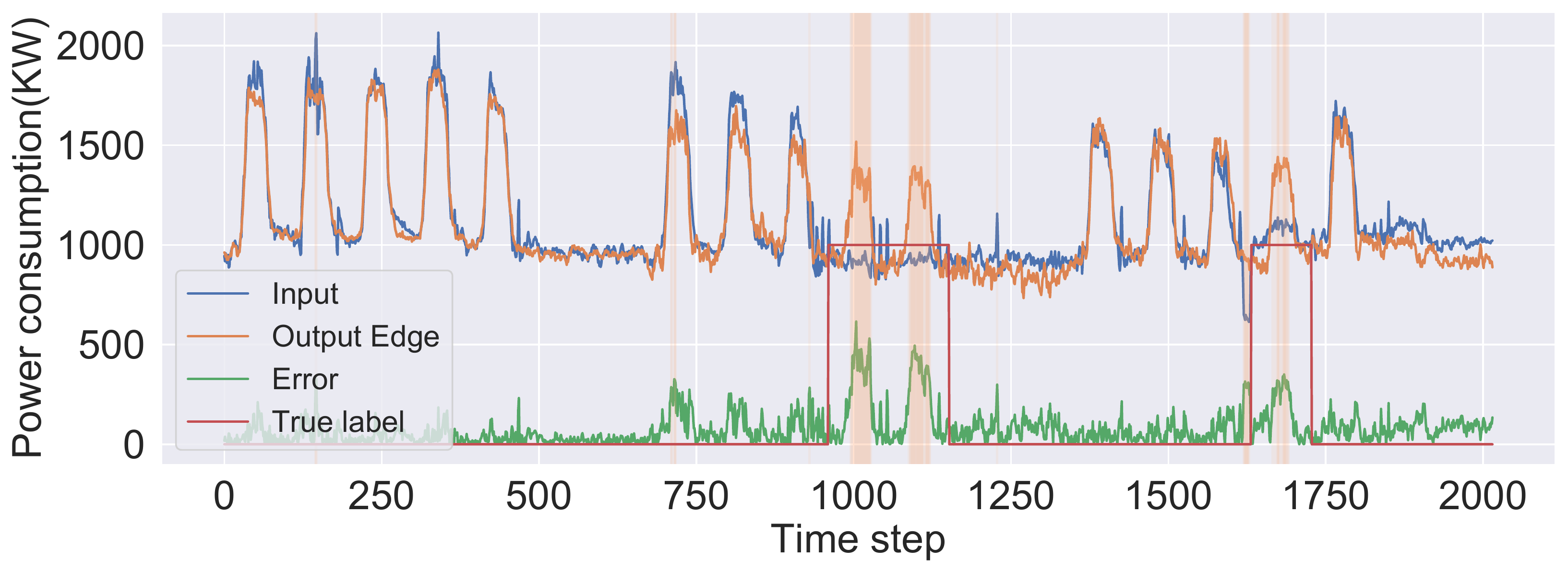}%
        \label{fig:TestReconstructionLossEdge}%
    } \quad
    \subfloat[Logarithmic probability densities (logPD) of errors.]{
        \includegraphics[width=1.0\linewidth]{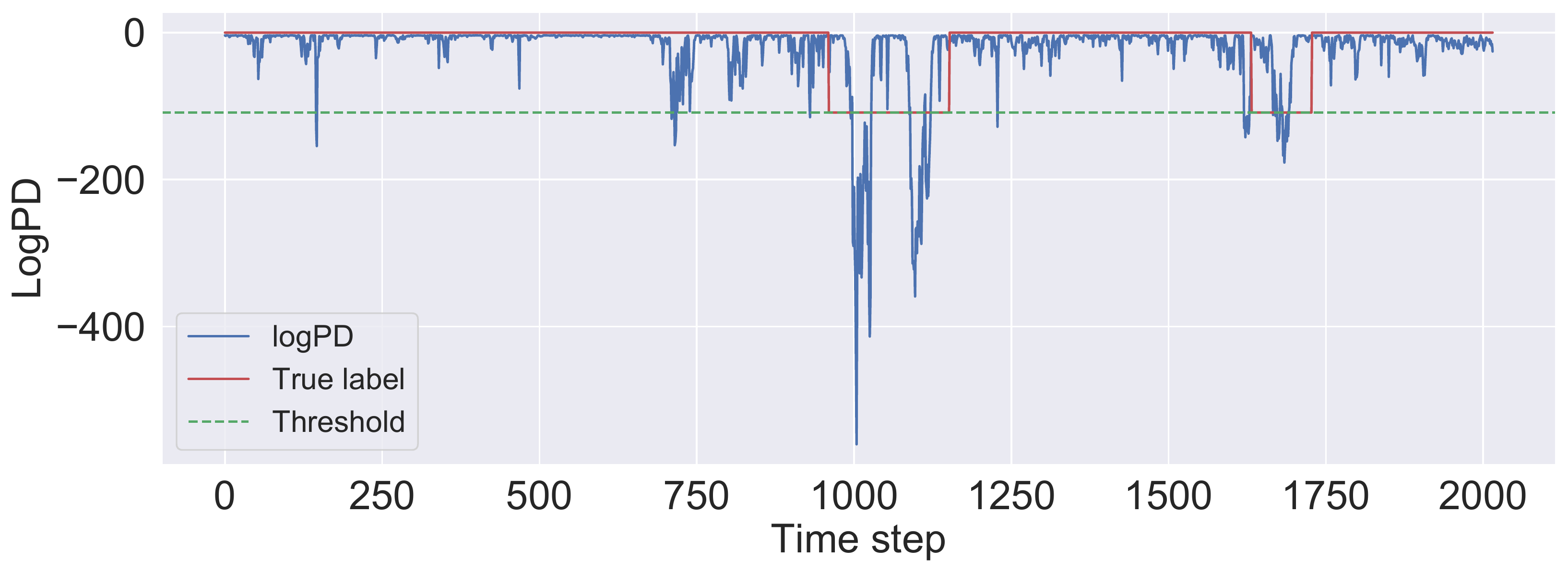}%
        \label{fig:TestLogPDEdge}%
    }
    \caption{AE-Edge performance on the first 2000 time steps of the test set for anomaly detection on edge servers.}
    \label{fig:ResultEdgeModel}
\end{figure}

\textbf{Anomaly detection models:}
The complexity of models increases from IoT to cloud, as indicated by the number of parameters and total FLOP of the models shown in Table \ref{tab:threeAnomalyDetectors}.
F1-score and accuracy of AE-IoT, AE-Edge, and AE-Cloud on the test set are obtained as (0.465, 78.09\%), (0.741, 93.33\%), and (0.870, 97.14\%), respectively.
We can see that the F1-score and accuracy of AE-Cloud are 87\% and 19\% higher than those of the AE-IoT model. 
Due to the space constraint, we show an example of reconstruction performance of the AE-Edge model on the first 2000 time steps of the test set in Fig.~\ref{fig:ResultEdgeModel}.

\textbf{Adaptive model selection scheme:}
The F1-score, accuracy, total reward, and average detection delay over the entire dataset under the five baseline schemes and our proposed scheme ($\alpha=0.0025$) are shown in Table \ref{tab:ExperimentResultDynamicScheme}.
The AE-IoT scheme achieves the smallest detection delay but the lowest accuracy and F1-score among all the evaluated schemes.
On another extreme, the AE-Cloud scheme provides the best accuracy but incurs the largest detection delay. 
The two Successive schemes leverage distributed anomaly detection models of HEC and thus reduce the average detection delay significantly as compared to AE-Edge and AE-Cloud.
However, their accuracy and F1-score are still below those of AE-Edge. 
In contrast, our proposed scheme adaptively chooses a suitable model to execute the anomaly detection task to jointly maximize accuracy and minimize detection delay.
Thus, not only does it has a smaller average detection delay, but its F1-score and accuracy also consistently outperform those of AE-IoT, AE-Edge, and the two Successive schemes. 
Even though the F1-score and accuracy of our proposed scheme are marginally (4\% and 0.28\%) lower than those of the AE-Cloud scheme, it reduces the average detection delay by a substantial 84\%.

In summary, our proposed adaptive anomaly detection scheme strikes the best tradeoff between accuracy and detection delay, and obtains the highest total reward. This is achieved by leveraging the HEC architecture and our policy network that automatically selecting the best layer to execute the detection task.

\begin{figure}[tb]
    \centering
    \includegraphics[width=1.0\linewidth]{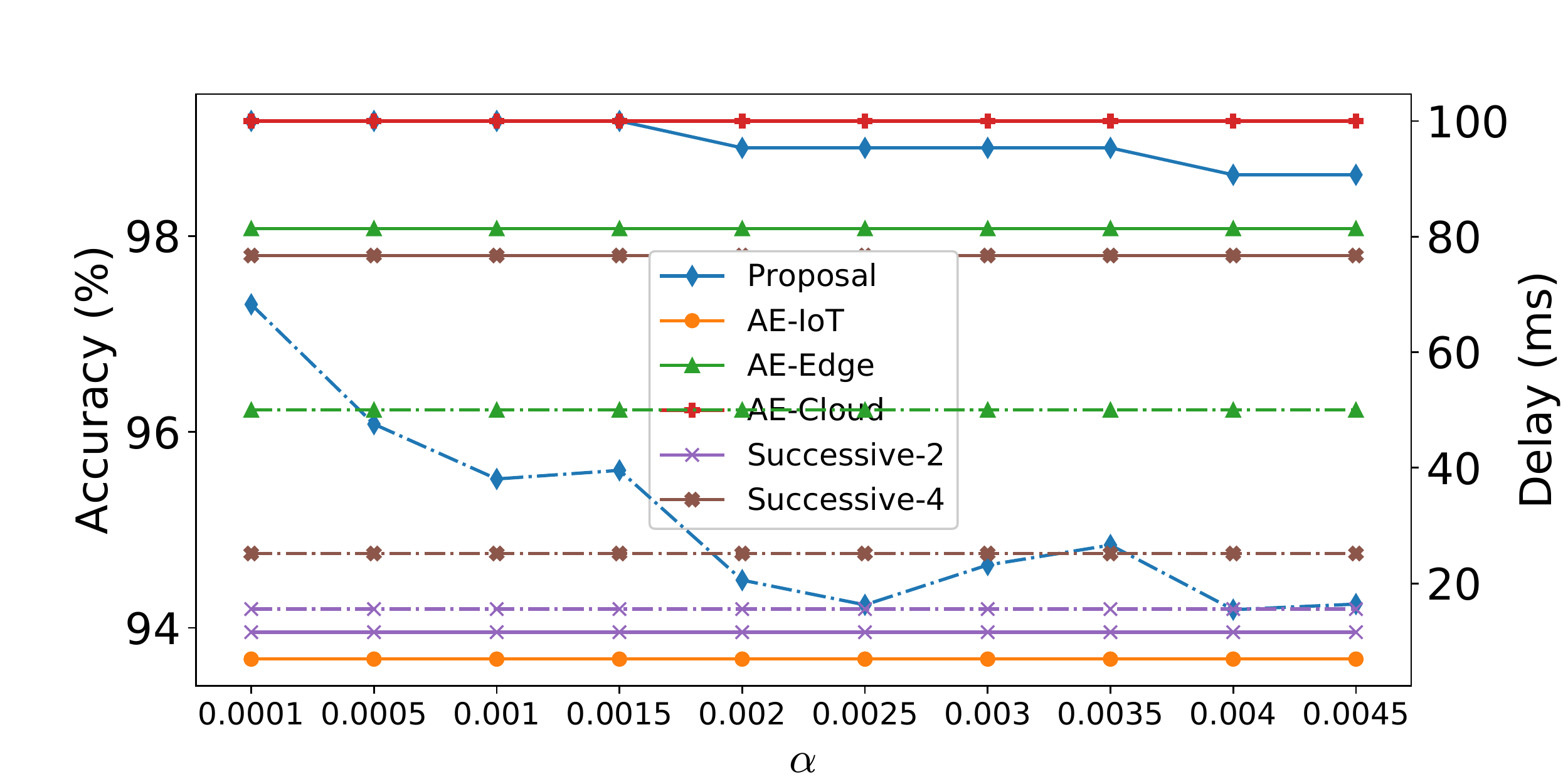}%
    \caption{Evaluation of how $\alpha$ affects our proposed scheme (through the cost function). Accuracy is plotted as solid lines while delay is plotted as dash lines.}
     \label{fig:tradeoffAccuracyDelay}
\end{figure}

\begin{table}[ht] 
    \caption{Comparison of Accuracy, Delay, and total Reward of all the anomaly detection schemes.}
    \label{tab:ExperimentResultDynamicScheme}
    \centering 
    \footnotesize 
    \begin{tabular}{ l@{\hspace{1em}} c@{\hspace{1em}}   c@{\hspace{1.em}}  c@{\hspace{1.em}}  c@{\hspace{1.em}} }
        \toprule
        {\textbf{Scheme}}   &  \textbf{F1} & \textbf{Accuracy(\%)} & \textbf{Reward} & \textbf{Delay(ms)}\\%
        \midrule%
        \textbf{AE-IoT}    & 0.465 & 93.68 & 47.18 & 6.89\\
        \textbf{AE-Edge}   & 0.741 & 98.08 & 41.64 & 50.00\\
        \textbf{AE-Cloud}  & 0.870 & 99.18 & 35.71 & 100.00\\
        \textbf{Successive-2} & 0.476 &  93.96 &   N/A    & 15.55\\
        \textbf{Successive-4} & 0.714 &  97.80 &   N/A    & 25.17\\
        \midrule%
        \textbf{Our Method}  & 0.833   & 98.90 & 49.52 & 16.28 \\
        \bottomrule
    \end{tabular}	
\end{table}

\textbf{Cost function to trade-off between accuracy and delay:}
We train different policy networks using different cost functions with the tunable parameter $\alpha\in [0.0001, 0.0045]$.
We plot the accuracy (solid line) and delay (dash line) of the proposed scheme and the baseline schemes (which are independent of $\alpha$) in Fig. \ref{fig:tradeoffAccuracyDelay}.
We can see that as $\alpha$ increases, both accuracy and delay of the proposed scheme generally decreases.
When $\alpha \leq 0.0015$, the proposed scheme achieves the highest accuracy as the AE-Cloud scheme, while reducing the average delay significantly by 40\%-60\%. Based on the overall performance, we choose $\alpha=0.0025$ for the best tradeoff between accuracy and delay.


\section{Conclusions}
\label{sec:conclusions}

In this paper, we identify three issues in existing IoT anomaly detection approaches, namely using one universal model to fit all data, lobesided focus on accuracy, and lack of local analysis. We then propose an adaptive anomaly detection scheme for IoT data in HEC. It constructs three anomaly detection models with increasing complexity, and associate them with three layers of HEC from bottom to top (i.e., IoT devices, edge servers, and cloud). 
Next, it uses an adaptive scheme that consists of a policy network as the solution to a single-step MDP or equivalently a contextual bandit problem, to select the best suited model based on the contextual information of input data.
The experiment results based on a real-world IoT dataset show that our proposed scheme outperforms other baseline schemes and achieves the best overall performance on accuracy and detection delay. 




\fontsize{9.0pt}{10.0pt}
\selectfont
\bibliography{references}
\bibliographystyle{aaai}


\end{document}